\documentclass[conference]{IEEEtran}
\IEEEoverridecommandlockouts
\usepackage{cite}
\usepackage{amsmath,amssymb,amsfonts}
\usepackage[linesnumbered,ruled,vlined]{algorithm2e}
\usepackage{graphicx}
\usepackage{textcomp}
\usepackage{xcolor}
\usepackage{subcaption}
\usepackage[hidelinks]{hyperref}
\usepackage{enumitem}
\usepackage{amsmath}
\usepackage{multirow} 
\usepackage{array}
\usepackage{url}
\usepackage{balance}

\newcommand{\PreserveBackslash}[1]{\let\temp=\\#1\let\\=\temp}
\newcolumntype{C}[1]{>{\PreserveBackslash\centering}p{#1}}

\def\BibTeX{{\rm B\kern-.05em{\sc i\kern-.025em b}\kern-.08em
    T\kern-.1667em\lower.7ex\hbox{E}\kern-.125emX}}
\begin{document}

\makeatletter
\newcommand{\linebreakand}{%
  \end{@IEEEauthorhalign}
  \hfill\mbox{}\par
  \mbox{}\hfill\begin{@IEEEauthorhalign}
}
\makeatother

\title{Model Predictive Control for Optimal Motion Planning of Unmanned Aerial Vehicles
\thanks{Video: {\tt\url{https://youtu.be/G_Sor9JHhlY}}}
}
\author{\IEEEauthorblockN{1\textsuperscript{st} Duy-Nam Bui}
\IEEEauthorblockA{\textit{Vietnam National University}\\
Hanoi, Vietnam \\
duynam@ieee.org}
\and
\IEEEauthorblockN{2\textsuperscript{nd} Thu Hang Khuat}
\IEEEauthorblockA{\textit{Vietnam National University}\\
Hanoi, Vietnam \\
23025115@vnu.edu.vn}
\and
\IEEEauthorblockN{3\textsuperscript{rd} Manh Duong Phung}
\IEEEauthorblockA{\textit{Fulbright University Vietnam}\\
Ho Chi Minh City, Vietnam \\
duong.phung@fulbright.edu.vn}
\linebreakand
\IEEEauthorblockN{4\textsuperscript{th} Thuan-Hoang Tran}
\IEEEauthorblockA{\textit{Center of Electrical Engineering} \\
\textit{Duy Tan University}\\
Da Nang, Vietnam \\
tranthuanhoang@duytan.edu.vn}
\and
\IEEEauthorblockN{5\textsuperscript{th} Dong LT Tran}
\IEEEauthorblockA{\textit{Center of Electrical Engineering} \\
\textit{Duy Tan University}\\
Da Nang, Vietnam \\
tranthangdong@duytan.edu.vn}
}






\maketitle

\begin{abstract}
Motion planning is an essential process for the navigation of unmanned aerial vehicles (UAVs) where they need to adapt to obstacles and different structures of their operating environment to reach the goal. This paper presents an optimal motion planner for UAVs operating in unknown complex environments. The motion planner receives point cloud data from a local range sensor and then converts it into a voxel grid representing the surrounding environment. A local trajectory guiding the UAV to the goal is then generated based on the voxel grid. This trajectory is further optimized using model predictive control (MPC) to enhance the safety, speed, and smoothness of UAV operation. The optimization is carried out via the definition of several cost functions and constraints, taking into account the UAV's dynamics and requirements. A number of simulations and comparisons with a state-of-the-art method have been conducted in a complex environment with many obstacles to evaluate the performance of our method. The results show that our method provides not only shorter and smoother trajectories but also faster and more stable speed profiles. It is also energy efficient making it suitable for various UAV applications.
\end{abstract}


\section{Introduction}
The ability of unmanned aerial vehicles (UAVs) to navigate in unknown environments, where traditional maps and pre-existing data are scarce or non-existent, is becoming more important due to their increasing roles in various applications. Whether deployed for search and rescue missions in remote areas or for exploration in complex terrains, UAVs need a robust navigation system to adapt to the uncertainties of their surroundings~\cite{PHUNG2020106705, PHUNG201725}. This system relies on advanced sensors such as GPS, Lidar, and RGB-D cameras to collect data about the environment and robust algorithms to process that data for real-time decision-making~\cite{8756125}. One of the key algorithms for navigation is motion planning, which is the process of determining a feasible trajectory for the UAV to move from its current location to a desired goal location while avoiding obstacles and adhering to various constraints. An optimal motion planning algorithm not only enhances efficiency but also ensures the safe and reliable operation of the UAV, making it essential for a UAV system. 

Early work on motion planning represents the environment via virtual forces within potential fields (PFs)~\cite{Garibeh2022, 8903321}. The method creates an artificial force field where obstacles repel the robot and the target attracts it. The field thus guides the robot toward the target while avoiding obstacles. In particular, a collision-free real-time motion planning method using potential force functions is introduced in~\cite{Garibeh2022} for UAVs to track dynamic targets and avoid multiple obstacles under low hover conditions. In~\cite{9234396}, a dynamic artificial potential field motion planning technique is introduced for multi-rotor UAVs to track moving targets using range sensors. This approach is simple to implement and can generate smooth paths. However, it requires the UAV to decelerate when approaching obstacles~\cite{131810} and hence reduces its operation efficiency, especially in environments with high obstacle density. It also has the local minimum problem, which causes the UAV to be trapped in an area with balanced virtual forces such as U-shaped corners.


In another approach, optimization and interpolation techniques have been used for motion planning. In~\cite{8202160}, a real-time approach for local trajectory planning for micro-vehicles capable of handling obstacles is introduced using B-spline to expand a local planning algorithm. In~\cite{8758904}, a motion planning system based on B-spline optimization is proposed for fast flight in complex three-dimensional space with trajectory smoothness enhanced. A cognitive-aware re-planning framework is presented in~\cite{9422918} to support fast and safe flight. It uses a path guidance optimization approach that combines multiple topological paths to find feasible routes in a short time. However, these methods consider geometric motion without considering the UAV's kinematic or dynamic constraints, making it infeasible for the UAV to track in certain circumstances.  

Model predictive control (MPC) can address the aforementioned issue due to its capability to solve the optimization problem subject to constraints, including physical constraints such as speed limits and control inputs, and environmental constraints~\cite{8424034, Song2023}. In particular, a Lyapunov-based nonlinear MPC is introduced in~\cite{Bui2022} to control a quadrotor to track a predefined trajectory in harsh conditions subject to noise and input limits. In~\cite{8594266}, a linear model predictive controller with nonlinear state feedback is proposed to track an optimal trajectory with high accuracy and collision avoidance capacity. However, current works still mainly use MPC for UAV control rather than motion planning due to its high computational cost in solving the objective equation~\cite{Darby2012, Bui2022}. Recent advancements in computational techniques and hardware capability can overcome this problem to extend MPC for motion planning~\cite{Bui2022, 2020SciPy-NMeth}.

In this study, we present an optimal motion planning method using MPC for UAVs in unknown complex environments. The UAV is equipped with a local sensor to provide point cloud data of the surrounding environment. This observed data is then converted to a voxel grid for local trajectory generation. This trajectory is optimized based on a set of cost functions designed to achieve the requirements for UAV motion. The optimization process also considers both the high-level dynamics and physical constraints of the UAV. Our contributions are threefold:
(i) propose a two-phase motion planner using real-time point cloud data observed from the environment by a local sensor equipped on the UAV;
(ii) define a set of cost functions to turn the motion planning into an optimization problem considering requirements and constraints for efficient operation of the UAV in dense environments;
(iii) solve the cost functions using MPC and validate its performance through various simulations and comparisons.
\section{System model}\label{sec:bg}
\begin{figure}
    \centering
    \includegraphics[width=0.45\textwidth]{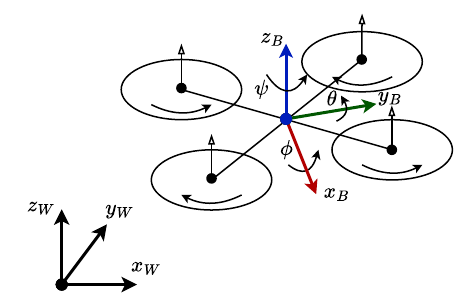}
    \caption{The quadrotor UAV model}
    \label{fig:scheme}
\end{figure}

The UAV used in this work is a quadcopter equipped with an inertial measurement unit (IMU) and a GPS module for positioning, and a range sensor for collecting point cloud data of the surrounding environment. The UAV state includes its position $p\in\mathbb{R}^3$, attitude $(\phi$, $\theta$, $\psi)$, velocity $v\in\mathbb{R}^3$, and acceleration $a\in\mathbb{R}^3$, as illustrated in Fig. \ref{fig:scheme}. The control inputs include attitude $\left(\phi_\text{cmd},\theta_\text{cmd},\psi_\text{cmd}\right)$ and thrust $T_\text{cmd}$. The forces acting on the multirotor are the gravity, drag forces, and thrust of the rotors. The dynamic equations of the quadcopter are given as follows \cite{Foehn2021,Furrer2016,9560773}:
\begin{equation}
    \begin{aligned}
        \dot{p}&=v\\
        \dot{v}&=-gz_W+\dfrac{T_\text{cmd}}{m}z_B-RDR^Tv\left\Vert v\right\Vert\\
        \dot{\phi}&=\dot{\phi}_\text{cmd}\\
        \dot{\theta}&=\dot{\theta}_\text{cmd}\\
        \dot{\psi}&=\dot{\psi}_\text{cmd}
    \end{aligned}
    \label{eq1}
\end{equation}
where $g$ is the gravitational acceleration, $m$ is the UAV's mass, $D\in\mathbb{R}^{3\times3}$ is the drag matrix, $R\in\mathbb{R}^{3\times3}$ is the rotation matrix from the body to the world frame, $z_B$ and $z_W$ are respectively the $z$ vectors of the body and the world frames. Equations in (\ref{eq1}) can be simplified into the following linear model with jerk $j$ as the input \cite{9560773,Toumieh2022}:

\begin{equation}
    \begin{aligned}
        &p(k+1)= p(k)+\tau\cdot v(k)\\
        &v(k+1)= v(k)+\tau\cdot\left(a(k)-D_\text{max}v(k)\right)\\
        &a(k+1)= a(k)+\tau\cdot j(k)\\
    \end{aligned}
    \label{eqn:dynamic}
\end{equation}
where $D_\text{max}\in\mathbb{R}^{3\times3}$ is a diagonal matrix representing maximum linear drag coefficients in all directions and can be identified offline as in \cite{9560773}. Denote  $x(k)=\left[p(k),v(k),a(k)\right]^T\in\mathbb{R}^9$ and $u(k)=j(k)\in\mathbb{R}^3$ respectively as the state and control input at the time $t(k) = k\tau$, with $\tau$ is the sampling period. The discrete dynamics of the UAV in \eqref{eqn:dynamic} can be rewritten in the matrix form as follows:
\begin{equation}
    x(k+1) = Ax(k) + Bu(k),
\end{equation}
where $A=\left[\begin{array}{ccc}
I_{3\times3} & \tau I_{3\times3} & 0_{3\times3}\\
0_{3\times3} & I_{3\times3}-\tau D_\text{max} & \tau I_{3\times3}\\
0_{3\times3} & 0_{3\times3} & I_{3\times3}
\end{array}\right]\in\mathbb{R}^{9\times9}$ and $B=\left[\begin{array}{c}
0_{3\times3}\\
0_{3\times3}\\
\tau I_{3\times3}
\end{array}\right]\in\mathbb{R}^{9\times3}$. This model is used to develop our motion planning method described in the next section.

\section{Motion planning method}\label{sec:me}
\begin{figure*}
    \centering
    \includegraphics[width=0.9\textwidth]{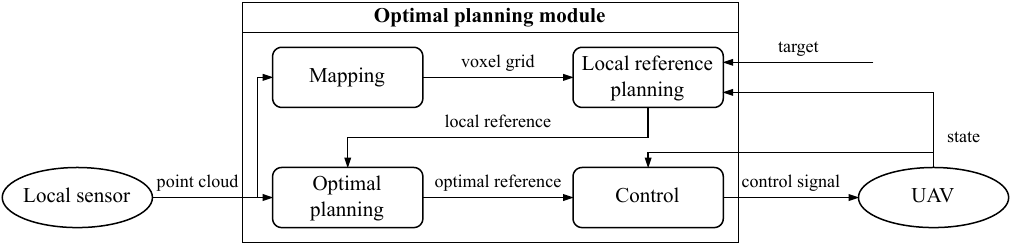}
    \caption{The proposed motion planning system}
    \label{fig:framework}
\end{figure*}
\begin{figure*}
    
    \centering
    \includegraphics[width=0.8\textwidth]{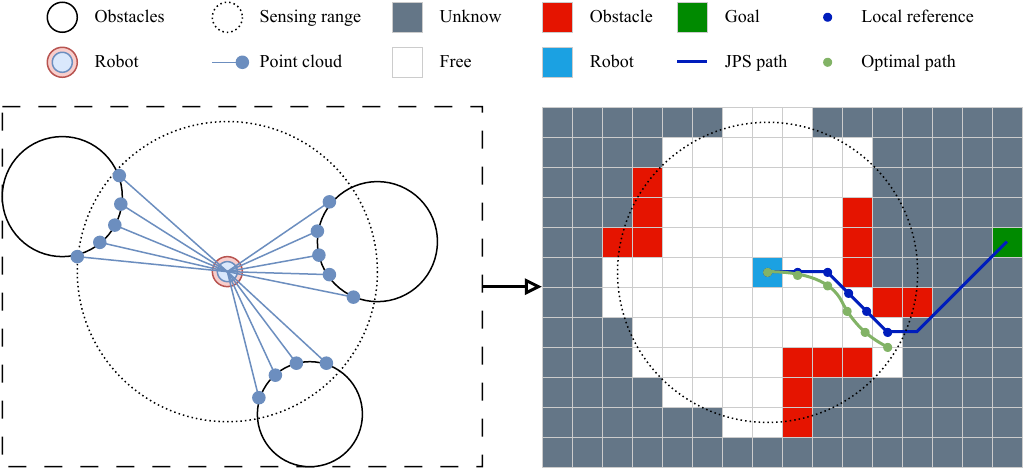}
    \caption{The proposed motion planning approach. Left: The point cloud data obtained from the range sensor. Right: The planning trajectories in the voxel grid, where the blue path is the global path generated by the JPS algorithm; the blue points are the local reference sampled from the global path; the green path is the local optimal trajectory generated by MPC.}
    \label{fig:method}
\end{figure*}

The motion planner aims to create an optimal local trajectory to navigate the UAV to the goal in densely clustered environments. Fig. \ref{fig:framework} presents the proposed motion planner with four main modules as follows:

\begin{enumerate}[label=(\roman*)]
    \item \textit{Mapping}: this module converts the point cloud data collected from the local range sensor to a voxel grid representing the surrounding environment of the UAV.  
    \item \textit{Local reference planning}: this module uses the voxel grid and the target information to generate a local reference trajectory.
    \item \textit{Optimal planning}: An optimal planning module refines the local reference trajectory to satisfy smoothness, safety, and velocity constraints.
    \item \textit{Control}: This module computes control signals $\left(T_\text{cmd},\phi_\text{cmd},\theta_\text{cmd},\psi_\text{cmd}\right)$ for the UAV to track a reference trajectory.
\end{enumerate} 

With those modules, the motion planner works as illustrated in Fig. \ref{fig:method}. When the sensing data is updated from the local sensor, a local map is created in the form of a voxel grid. A global shortest path is then generated using the Jump Point Search (JPS) algorithm \cite{Harabor2011}. A local reference trajectory $p_\text{ref}$ is then generated, comprising $P$ points sampled from the global path at a reference velocity $v_\text{ref}$. The final trajectory is then obtained by minimizing a cost function. The function is a weighted sum of sub-cost functions including the trajectory tracking $J_t$, speed $J_s$, collision $J_c$, and jerk $J_j$ terms. These functions are defined as follows.

\textit{Tracking cost:}
The tracking term $J_t$ is used to minimize the difference between the local reference path and the generating path. It is defined as follows:
\begin{equation}
    J_{t}(k)=w_t\sum_{i=1}^P{\left\Vert p(k+i|k)-p_\text{ref}(k+i|k)\right\Vert^2},
\end{equation}
where $w_t$ is a positive tracking weight.

\textit{Speed cost:}
The speed cost $J_s$ aims to maintain the desired flight speed $v_\text{ref}$. It is defined as
follows:
\begin{equation}
    J_{s}(k)=w_s\sum_{i=1}^P\left(\left\Vert v(k+i|k)\right\Vert^2 - v_\text{ref}^2\right)^2,
\end{equation}
where $w_s$ is the positive speed weight.

\textit{Collision cost:}
The collision cost $J_c$ is created to avoid collision between the UAV and obstacles. It is defined based on the logistic function \cite{8202163} as follows:

\begin{equation}
    J_{c}(k)=w_c\sum_{i=1}^P\sum_{m\in\mathcal{M}} \dfrac{1}{1 + \exp{\left(\alpha\left(d_m(k+i|k) - r\right)\right)}},
\end{equation}
where $d_m$ is the Euclidean distance from the UAV to obstacle $m$, $d_m(k+i|k)= \left\Vert p(k+i|k) - p_m\right\Vert^2$; $\alpha > 0$ is a parameter representing the smoothness of the cost function; and $r>0$ is a pre-defined safety distance.

\begin{figure*}
    \centering
    \includegraphics[width=0.9\textwidth]{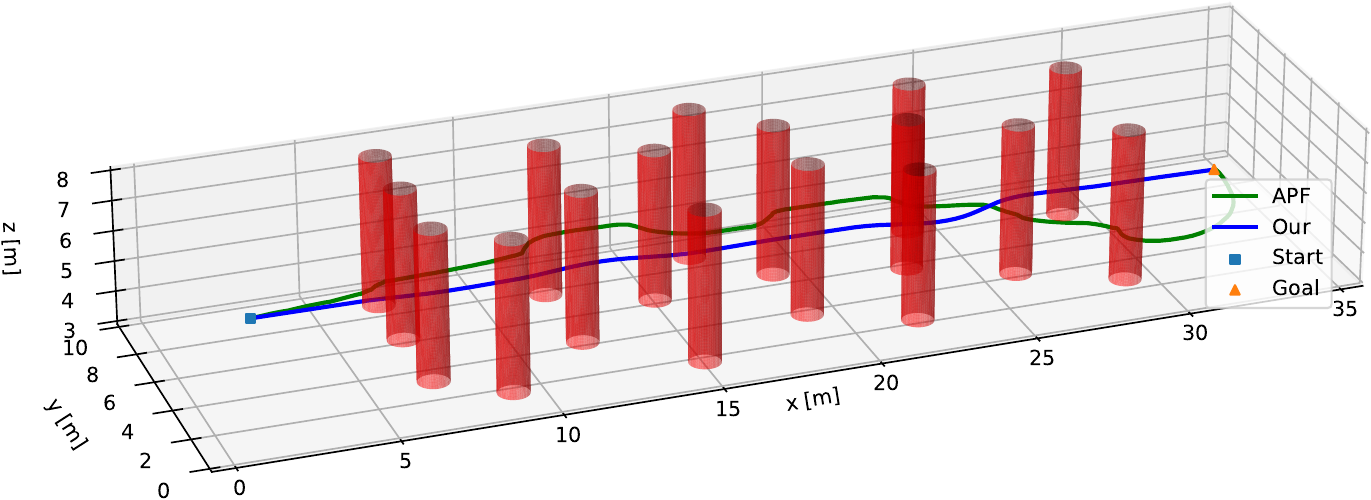}
    \caption{3D view of the generated trajectories.}
    \label{fig:3d}
\end{figure*}

\textit{Jerk penalty:}
The control cost is used as a penalty term to obtain a minimal and smooth jerk. It is given as follows:
\begin{equation}
    J_{j}(k)=w_j\sum_{i=1}^P\left\Vert u(k+i|k)\right\Vert^2,
\end{equation}
where $w_j$ is a positive weight.

Let $X(k)\in\mathbb{R}^{9P}$ be the sequence of states $x(k+i|k)$ at points $i\in\{1,...,P\}$ and $U(k)\in\mathbb{R}^{3P}$ be the sequence of control inputs $u(k)$ over the horizon $i\in\{0,...,P-1\}$. The optimal trajectory can be obtained through the following non-convex optimization: 
\begin{equation}
    \min_{U(k),X(k)} \left(J_t + J_s + J_c + J_j\right)
    \label{eq:optimization}
\end{equation}
subject to
\begin{equation}
\label{eq:optimization2}
    \begin{aligned}
        & x(k+i+1|k)=Ax(k+i|k)+Bu(k+i|k)\\
        & \left\Vert v(k+i|k)\right\Vert\leq v_\text{max}\\
        & \left\Vert a(k+i|k)\right\Vert\leq a_\text{max}\\
        & \left\Vert u(k+i|k)\right\Vert\leq u_{\text{max}}
    \end{aligned}
\end{equation}
with $i\in\left\{1,...,P\right\}$. Equation (\ref{eq:optimization2}) ensures that the generated trajectory meets the constraints imposed by physical limits in velocities and control inputs of the UAV. In addition, the jerk cost is used to maintain the smoothness of the generated trajectory.


\begin{algorithm}[!ht]
\caption{Pseudocode of generating local trajectory $X(k)$ using MPC}
\label{alg:mpc_sqp}
\SetAlgoLined
\tcc{Initialization}
Set initial value for control trajectory $U(k-1)$\;
Set initial state $x(k|k)=x(k)$\;
Set prediction horizon $P$\;
Set cost function $J$\;
Set constraints on control inputs and states\;

\tcc{Main Loop}
\While{not converged}{
    \For{$i \leftarrow 0$ \KwTo $P-1$}{
        Predict $x(k+i+1|k)$\tcc*[r]{Eq. \ref{eq:optimization2}}
    }
    Formulate the cost function $J(k)$\tcc*[r]{Eq. \ref{eq:optimization}}
    Define inequality constraints\tcc*[r]{Eq. \ref{eq:optimization2}}
    
    Linearize the dynamics and constraints around the current trajectory and solve the quadratic programming subproblem to get the control trajectory $U(k)$\tcc*[r]{Ref. \cite{kraft1988software}}
    
    Update state trajectory $X(k)$\tcc*[r]{Eq. \ref{eq:optimization2}}
    
    \If{convergence criteria is met}{
        break\;
    }
}

\Return $X(k)$\;
\end{algorithm}

We use MPC to solve the optimization \eqref{eq:optimization} - \eqref{eq:optimization2} to obtain the local trajectory $X(k)$. The system dynamics and the constraints of the problem are discretized over the prediction horizon to obtain a structured nonlinear program (NLP). Sequential least squares programming (SLSQP)~\cite{kraft1988software} is used to produce the optimal solution based on its cost function. Algorithm \ref{alg:mpc_sqp} describes the MPC solver using SLSQP method to generate the local reference $X(k)$. In practical, we implemented the MPC solver in Python using SciPy library \cite{2020SciPy-NMeth} to enhance computational speed.

\begin{table}[!ht]
\centering
\caption{Planners comparison}
\label{tbl:com}
\begin{tabular}{C{1.5cm}C{1.8cm}C{2.2cm}C{1.5cm}} 
\hline \hline
Planner & Motion time (s)   & Motion length (m) & Energy \\ \hline
APF     & 70.8              & 38.4481               & 568.0819 \\
Our method     & \textbf{33.3}     & \textbf{31.2679}  & \textbf{169.9341} \\
\hline \hline
\end{tabular}
\end{table}

\begin{figure*}[!ht]
    \centering
    \begin{subfigure}[b]{0.9\textwidth}
    \includegraphics[width=\textwidth]{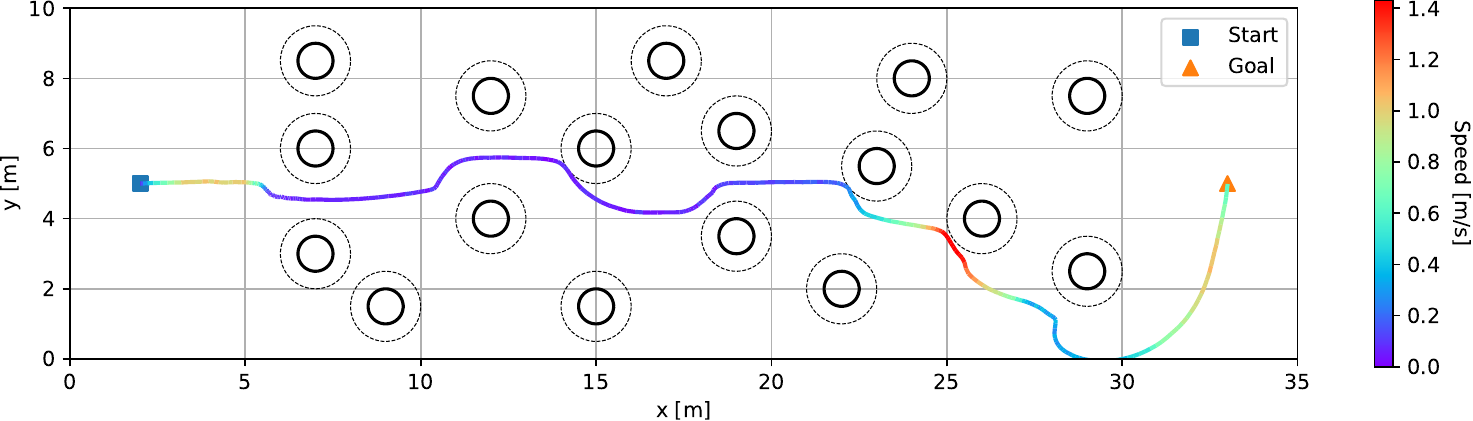}
    \caption{The APF planner}
    \label{fig:apf}
    \end{subfigure}
    \begin{subfigure}[b]{0.9\textwidth}
    \includegraphics[width=\textwidth]{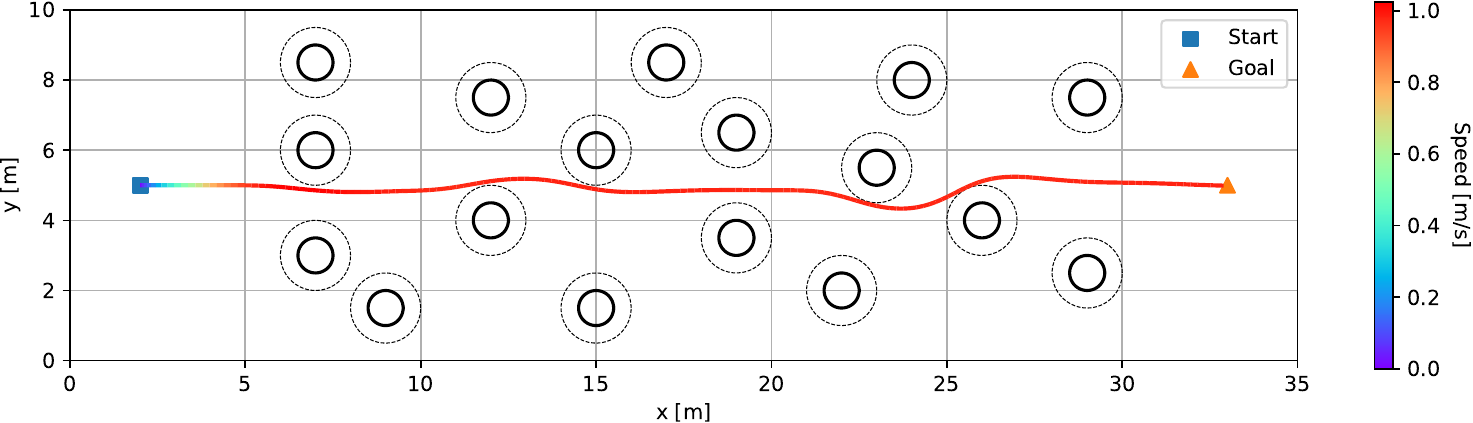}
    \caption{The proposed planner}
    \label{fig:our}
    \end{subfigure}
    \caption{Top view of the generated trajectories together with their speed profiles. The solid and dashed black lines, respectively, represent the obstacles in the environment and their extent to accommodate the robot's size.}
    \label{fig:top}
\end{figure*}

\begin{figure*}[!ht]
    \centering
    \begin{subfigure}[b]{0.28\textwidth}
    \frame{\includegraphics[width=\textwidth]{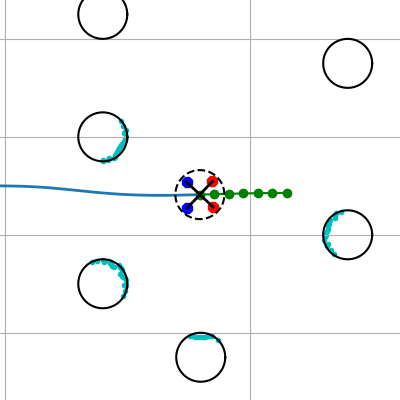}}
    \caption{$t=9.1$ s}
    \end{subfigure}
    \hspace{0.5cm}
    \begin{subfigure}[b]{0.28\textwidth}
    \frame{\includegraphics[width=\textwidth]{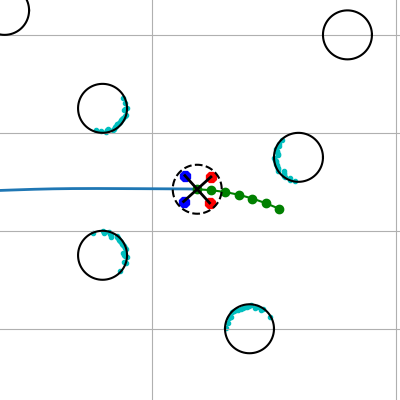}}
    \caption{$t=21.8$ s}
    \end{subfigure}
    \hspace{0.5cm}
    \begin{subfigure}[b]{0.28\textwidth}
    \frame{\includegraphics[width=\textwidth]{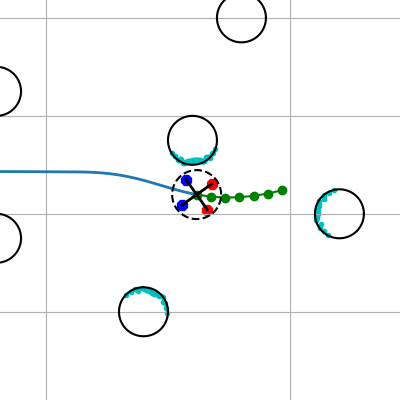}}
    \caption{$t=23.3$ s}
    \end{subfigure}
    \caption{Snapshots of the proposed planner: the blue line shows the motion path of the UAV; the green points show the local trajectory generated at each time step; the black circles represent obstacles in the environment; and the cyan points are the point cloud generated by the local sensor.}
    \label{fig:motion}
\end{figure*}

\section{Results and Discussion}\label{sec:re}
To evaluate the performance of the proposed approach, we have conducted a number of simulations and comparisons with details as follows. 

\subsection{System setup}
The UAV has a size of 0.5~m and is equipped with a local range sensor with a sensing range of 3~m. It has the maximum velocity $v_\text{max}=2.0$ m/s, maximum acceleration $a_\text{max}=g = 9.81$ m/s$^2$, maximum jerk $u_\text{max}=j_\text{max}=1.0$ m/s$^3$, and maximum drag coefficients $D_\text{max}=\text{diag}(0.5,0.5,0.5)$. The desired velocity $v_\text{ref}$ is set to 1.0~m/s. The number of trajectory points is set to $P=20$. The control sampling period is $\tau=0.1$~s. 

Comparisons with the artificial potential field (APF) planner~\cite{Garibeh2022, 9234396} are conducted for evaluation. The comparison metrics used include motion time, motion length, and energy consumption. The energy consumption is measured by the acceleration integral method~\cite{Zhang2021}. 

\subsection{Results}
Fig. \ref{fig:3d} shows the 3D view of the trajectories generated by the two algorithms and Fig. \ref{fig:top} presents their top view with the speed profiles. It can be seen that both planners successfully navigate through the dense obstacle environment. However, the trajectory generated by the APF does not maintain the desired speed $v_\text{ref}$. Instead, it slows the UAV down when approaching obstacles leading to higher motion time. In contrast, the proposed approach provides a shorter and smoother trajectory. It also maintains a higher and more stable speed profile. This result can be further confirmed in Table \ref{tbl:com} which shows the performance of both planners. It can be seen that the proposed method outperforms the APF in all metrics with two times less motion time and three times less energy consumption. Fig. \ref{fig:motion} shows several snapshots of the proposed planner in the environment. Through the point cloud observed by the local sensor (cyan points), the proposed planner can provide the optimal local trajectory (green points) that satisfies the constraints on safety and speed while minimizing the jerk.
\section{Conclusion}\label{sec:cl}
In this paper, we have presented an optimal motion planner that can safely and smoothly navigate a UAV through dense obstacle environments. The planner can process the point cloud data from a local sensor in real-time to generate optimal local trajectories. By modeling the motion planning as an optimization problem, the output of the planner is an optimal local trajectory that, at each period, satisfies the constraints on safety, speed, and smoothness. Simulations and comparisons confirm the superiority of the proposed method compared to state-of-the-art methods such as the APF.

\section*{Acknowledgement}
Thu Hang Khuat was funded by the Master, PhD Scholarship Programme of Vingroup Innovation Foundation (VINIF), code VINIF.2023.Ths.044.

\balance
\bibliographystyle{ieeetr}
\bibliography{ref}
\end{document}